\documentclass[10pt, a4paper, twocolumn]{naverlabseurope} 

\usepackage{lipsum}
\usepackage{multicol}
\usepackage{tikz,tkz-kiviat,pgfplots}

\usepackage[utf8]{inputenc}
\usepackage{amsmath}
\usepackage{amsfonts}
\usepackage{amssymb}
\usepackage{url}
\usepackage{booktabs}
\usepackage{hyperref}
\usepackage{microtype}
\usepackage{wrapfig}
\usepackage{float}
\usepackage{graphicx}
\usepackage{algorithm2e}
\usepackage[noend]{algpseudocode}
\usepackage[table]{xcolor} 
\usepackage{xstring}      
\usepackage{breqn}
\usepackage{caption} 

\usepackage{caption} 

\providecommand{\website}[1]{}
\renewcommand{\website}[1]{%
  \href{#1}{\textcolor{black}{#1}}%
}

\newcommand{\hlcell}[1]{%
  \IfEqCase{#1}{%
    {best}   {\cellcolor[HTML]{7CB696}}
    {second} {\cellcolor[HTML]{BEDBBD}}
    {third}  {\cellcolor[HTML]{E8F1E4}}
  }[\PackageWarning{hlcell}{Unknown value `#1`}]%
}

\graphicspath{{img/}}

\title{ProgVLA: Progress-Aware Robot Manipulation Skill Learning}
\titlerunning{ProgVLA}

\correspondingauthor{seungsu.kim@naverlabs.com}

\authors{
Seungsu Kim$^{2}$,
Jinyoung Choi$^{1}$,
Seungmin Baek$^{1}$,
Jean-Michel Renders$^{2}$
}

\affiliations{
$^{1}$ NAVER LABS \\
$^{2}$ NAVER LABS Europe
}

\begin{abstract}
We present ProgVLA, a compact vision-language-action (VLA) model designed for reliable robot manipulation under tight compute and memory budgets. The model specifically focuses on efficiently processing long multi-modal sequences by maintaining an explicit representation of task progress over extended horizons. To this end, ProgVLA integrates two key components.
First, a multi-modal encoder with a two-stage Perceiver resampling scheme compresses variable-length visual, language, and proprioceptive streams into a fixed set of control-ready context tokens, substantially reducing sequence length while preserving cross-modal grounding. Second, an auxiliary set of progress heads is trained with offline reinforcement learning (RL) objectives to jointly learn critics over normalized remaining-horizon targets. This provides the policy with an internal estimate of task progress and enables advantage- and success-weighted flow-matching imitation learning.
On two well-established multi-task robot manipulation benchmarks, a 0.1B-parameter ProgVLA model reaches success rates that are competitive with, and on long-horizon and harder task tiers exceed, substantially larger pretrained baselines. Ablations indicate that the learned context resampler and task-adaptive visual fine-tuning are the largest single contributors, while progress-aware training provides a  consistent additional gain that is concentrated on long-horizon and multi-object tasks. We further validate the approach in real-world toy-kitchen environments.
\end{abstract}

\begin{document}

\maketitle

\section{Introduction}
Vision-Language-Action (VLA) models aim to execute natural-language instructions by grounding perception in precise robotic manipulation. State-of-the-art VLAs pair multi-billion-parameter backbones with large-scale robot pretraining on cross-embodiment corpora~\cite{brohan_rt-2_2023,kim_openvla_2024,black_pi_0_2024,black_pi_05_2025}, limiting where and by whom they can be deployed. Small-footprint VLAs such as SmolVLA~\cite{shukor_smolvla_2025} run on modest hardware, but still rely on cross-embodiment pretraining and struggle on long-horizon tasks. We ask whether a single \emph{compact} model, trained from benchmark-scale demonstrations alone, can match or exceed larger pretrained baselines, focusing on the long-horizon regime where compact models lag the most.

\begin{figure*}[ht]
  \centering
  \includegraphics[width=13cm]{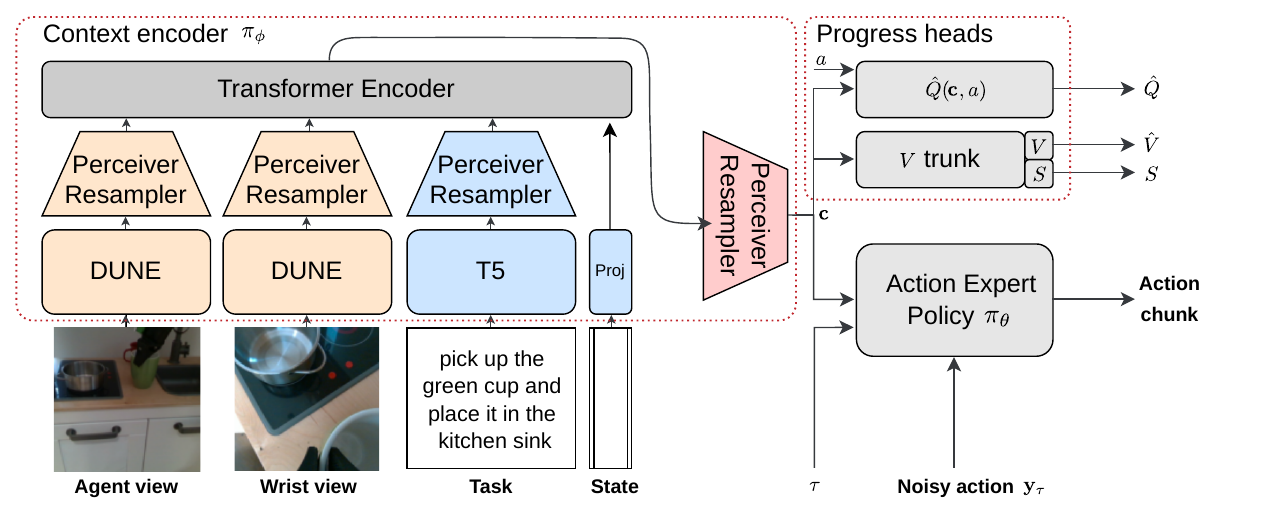}
  \caption{ProgVLA architecture. Per-modality Perceiver resamplers compress vision and language features to fixed-size token sets; a fusion Transformer and a post-fusion resampler then distill these into a small set of control-ready context tokens that condition both the flow-matching action expert and the auxiliary progress heads.}
  \label{fig:architecture}
\end{figure*}

We identify three complementary design requirements: (i) efficient cross-modal feature extraction under bounded compute; (ii) strong visual priors from a universal pre-trained backbone, in lieu of robot-data pretraining; and (iii) explicit temporal grounding that lets the policy estimate \emph{how far along} it is in the task. ProgVLA (Progress-aware VLA) realizes all three in a single 0.1B-parameter model (Fig.~\ref{fig:architecture}).

\paragraph{Two-stage Perceiver resampling.} Two Perceiver Resampler~\cite{alayrac_flamingo_2022,jaegle_perceiver_2021} stages---per-modality and post-fusion---compress multi-modal observations into a small fixed-size set of control-ready tokens, reducing sequence length by an order of magnitude. Our ablations show this is the single largest contributor to long-horizon success.

\paragraph{Universal vision backbone.} In lieu of robot-data pretraining, we use DUNE~\cite{sariyildiz_dune_2025}, a ViT-Small distilled from specialist vision models on diverse 2D and 3D tasks, which yields broader priors than single-source pre-training~\cite{oquab_dinov2_2024}. Fine-tuning DUNE jointly with the policy is the second-largest performance lever.

\paragraph{Auxiliary progress heads.} Lightweight heads regress a shaped remaining-horizon signal and a near-completion success indicator from the same context tokens used by the policy; their detached predictions advantage- and success-weight the imitation loss. Unlike prior progress estimators that operate post-hoc as separate networks~\cite{du_vision-language_2023,alakuijala_video-language_2024,ma_vision_2025}, this signal is consumed \emph{internally} by the same controller that produces actions.  

Taken together, these components produce a compact controller with broad visual priors, an aggressively compressed multi-modal context, and an internal progress signal that reweights imitation learning. ProgVLA uses \emph{no large-scale robot pretraining}; the only pretrained components are the off-the-shelf DUNE vision backbone and the frozen T5 text encoder.
\section{Related work}

\subsection{Small-scale VLA models for robot manipulation}
Large VLA models~\cite{brohan_rt-2_2023,black_pi_0_2024,black_pi_05_2025,kim_openvla_2024} have shown that internet-scale vision-language pretraining drives broad generalization and emergent semantic reasoning in a single end-to-end model, but such models remain massive and compute-hungry. SmolVLA~\cite{shukor_smolvla_2025} showed that an efficient 450M-parameter VLA can match models ten times larger when trained on the same cross-embodiment robot corpora, lowering the barrier to entry for robotic learning. ProgVLA pushes this trend further along two axes simultaneously: we shrink the deployed model by another factor of $2$--$4$ to 0.1B parameters, while training \emph{without reliance on large-scale robot pretraining}, using only the target benchmark's demonstrations..

Efficiency depends not only on parameter count but also on sequence length, motivating work on token compression. The Perceiver family~\cite{alayrac_flamingo_2022,jaegle_perceiver_2021} replaces exhaustive self-attention with cross-attention from a fixed set of learned latents, reducing attention complexity from quadratic to linear. Flamingo's Perceiver Resampler compresses arbitrary image features into a small set of visual tokens for a frozen language model, and RoboFlamingo reuses this OpenFlamingo backbone with a policy head for language-conditioned manipulation~\cite{li_vision-language_2024}. We adopt a related strategy but apply two stages of latent bottlenecks (per-modality and post-fusion) and combine them with task-adaptive fine-tuning of the vision encoder; ablations show that this combination is what produces the largest long-horizon gains. Unlike previous compact VLAs that focus primarily on parameter count and throughput, ProgVLA additionally targets the temporal dimension of long-horizon tasks.

\subsection{Self-monitoring and progress estimation}
For long-horizon manipulation, a robot must monitor its progress (e.g., \emph{``How far along am I?''}, \emph{``Have I finished?''}). Recent work casts success or progress estimation as vision-language reasoning. SuccessVQA~\cite{du_vision-language_2023} fine-tunes Flamingo to decide whether a task succeeded from an image or video and the instruction. Luo et al.~\cite{luo_vision-language_2024} adapt MiniGPT-4 as a binary success classifier on real robot trajectories. Alakuijala et al.~\cite{alakuijala_video-language_2024} train a Video-Language Critic that ranks video snippets by progress toward a language goal via cross-embodiment temporal contrastive learning. Generative Value Learning~\cite{ma_vision_2025} prompts a vision-language model to estimate progress in-context by ordering shuffled trajectory frames. Contrastive $\lambda$-Repformer~\cite{goko_task_2024} aligns ``before'' and ``after'' images with the textual instruction to detect object state changes.

Our work differs in two ways. First, prior approaches typically rely on an \emph{external} success classifier or a \emph{separate} value/reward network that observes the agent's trajectory post-hoc and is then used for evaluation, reward shaping, or downstream policy improvement. ProgVLA's auxiliary progress objectives share the policy's own context representation and are trained jointly with the flow-matching objective; their scalar predictions are detached and used as multiplicative weights on the imitation loss, so the progress signal acts as both an auxiliary representation regularizer and a per-sample reweighting term, while the deployed model remains a single compact policy. Second, prior work mostly attaches progress estimation to large pretrained vision-language models; we show that a small progress head, trained jointly with a flow-matching policy on a purely temporal target, suffices to deliver consistent long-horizon gains in a 0.1B-parameter setting. To our knowledge, this internal, policy-coupled progress-aware training has not previously been explored among small-scale VLAs.
\section{ProgVLA}
\label{sec:progvla}
ProgVLA has three components: (i) a multimodal encoder that compresses robot observations and the task instruction into a small set of context tokens, (ii) progress heads that score these context tokens for trajectory progress and proximity to completion, and (iii) a flow-matching action expert that generates robot actions from the same context. Figure~\ref{fig:architecture} gives an architectural overview.

We use the following notation. The time-step index is $t$, and $\mathbf{I}_t = [I^1_t, \ldots, I^C_t]$ denotes RGB images from $C$ cameras. The vector $\mathbf{q}_t$ collects proprioception (joint angles, joint velocities, gripper status). The language instruction for the $n^{\text{th}}$ episode is $l_n$, and the action chunk over horizon $H$ is $\mathbf{A}_t = [a_t, \ldots, a_{t+H-1}]$. The dataset $\mathcal{D}=\{ \{\mathbf{I}_t, \mathbf{q}_t \}_{t=1}^{T_n}, l_n \}_{n=1}^N$ contains $N$ episodes of length $T_n$. We write $\pi_{\phi}(\mathbf{c}_t \mid \mathbf{I}_t, \mathbf{q}_t, l_n)$ for the context encoder, which produces context $\mathbf{c}_t$, and $\pi_{\theta}(\mathbf{A}_t \mid \mathbf{c}_t)$ for the action expert.

\subsection{Multi-modal encoder}
\label{sec:mm_encoder}
The encoder follows a four-stage design: (i) modality-specific encoders extract features from vision $\mathbf{I}$, language $l$, and proprioception $\mathbf{q}$; (ii) lightweight per-modality resamplers convert these features into a fixed number of tokens; (iii) a shared Transformer performs cross-modal fusion over the concatenated token sets; and (iv) a final post-fusion resampler distills the fused representation into compact, control-ready context tokens.

\paragraph{Modality-specific encoders.} For vision, we use DUNE~\cite{sariyildiz_dune_2025}, a universal encoder distilled from multiple foundation models that attains competitive results on core computer-vision benchmarks. Language instructions are embedded with the frozen T5 text encoder~\cite{raffel_exploring_2020}, chosen for its robust handling of diverse vocabularies. Proprioceptive signals are mapped into the shared feature space via a simple MLP projection.

\paragraph{Two-stage Perceiver resampling.} We use a Perceiver Resampler~\cite{alayrac_flamingo_2022}, which employs a small set of learnable latent queries that are repeatedly updated through cross-attention blocks over the input sequence, yielding a compact fixed-size token set. We first apply it as a \emph{per-modality} resampler that normalizes each encoder's output (DUNE vision features and T5 language features) into a fixed number of tokens, ensuring balanced capacity before fusion. A shared Transformer then performs cross-modal self-attention over the concatenated modality tokens, and a second \emph{post-fusion} resampler distills the fused sequence into a compact set of context tokens $\mathbf{c}_t$. This two-stage design provides \emph{learned bottlenecks} that stabilize training across heterogeneous sequence lengths and expose a consistent interface to downstream control heads while preserving strong cross-modal conditioning.

\subsection{Action expert}
\label{sec:action-expert}
We adopt the compact flow-matching action expert introduced in SmolVLA~\cite{shukor_smolvla_2025}. Conditioned on the context $\mathbf{c}$ (Sec.~\ref{sec:mm_encoder}) and a ground-truth action chunk $\mathbf{A}$, we sample noise $\boldsymbol{\epsilon}=\mathbf{y}_{0}\sim\mathcal{N}(\mathbf{0},\mathbf{I})$ and step $\tau \sim \mathcal{U}[0,1]$, form the interpolant $\mathbf{y}_\tau = (1-\tau)\mathbf{y}_0 + \tau\mathbf{y}_1$ with $\mathbf{y}_1 \sim \mathbf{A}$, and train $\pi_\theta$ to predict the velocity transporting $\mathbf{y}_\tau$ toward the data:
\begin{equation}
\label{eq:flow-matching-loss}
\mathcal{L}_{\text{FM}} = \mathbb{E}_{\tau,\boldsymbol{\epsilon}} \left[ \left\| \pi_\theta(\mathbf{y}_\tau; \tau, \mathbf{c})- ( \mathbf{y}_1-\boldsymbol{\epsilon}) \right \|_2^2\right].
\end{equation}
At inference time we draw $\mathbf{y}_0 \sim \mathcal{N}(\mathbf{0},\mathbf{I})$ and integrate the learned flow ODE with $K=10$ Heun (explicit trapezoidal) steps conditioned on $\mathbf{c}$ to obtain $\hat{\mathbf{A}}$, execute the first few actions (receding-horizon), and replan with fresh observations. During training, the time variable $\tau$ is sampled from $\tau \sim \mathrm{Beta}(\alpha=2,\,\beta=2)$, following~\cite{shukor_smolvla_2025}.

\subsection{Progress heads}
\label{sec:progress-heads}
We complement the flow-matching imitation objective with auxiliary heads that supply (i) additional supervision on the shared context representation $\mathbf{c}_t$ and (ii) a scalar \emph{progress-derived reweighting} signal for the imitation loss. The design is inspired by offline reinforcement learning, but we deliberately call these objectives \emph{auxiliary}: we use only successful demonstrations, regress all value heads on the same Monte-Carlo target (no TD bootstrapping), and feed the resulting scalars back into training only as \emph{detached} per-sample weights; the policy itself is never updated by a policy-gradient signal.

\paragraph{Progress target.} We assume each dataset trajectory is successful, i.e., it terminates in a state satisfying the accompanying instruction. Let $T_n$ denote the terminal time index of episode $n$. Dropping the episode index for clarity, we define a normalized progress target
\begin{equation}
r_t = \max\!\left(0,\; 1 - \frac{T-t}{T_M}\right),
\end{equation}
where $T_M$ is a fixed horizon cap ($T_M = 500$ in our experiments). The signal $r_t \in [0,1]$ is a \emph{shaped progress signal} derived purely from trajectory phase rather than from a semantic subgoal detector or a raw environment reward, and serves as the regression target for both heads below. We discuss the implications of this purely temporal shaping in Sec.~\ref{sec:conclusion}.

\paragraph{Architecture of the heads.} As illustrated in Fig.~\ref{fig:architecture}, the context tokens $\mathbf{c}_t$ are fed into two trunks: (i) a \emph{state-action critic} $\hat Q(\mathbf{c}_t, a_t)$ that projects $a_t$ through an MLP into a single token, concatenates it with the context tokens, and pools through a small Perceiver pooler; and (ii) a \emph{value trunk} that pools the context tokens alone and feeds the pooled feature into two parallel scalar heads, producing the progress value $\hat V(\mathbf{c}_t)$ and the success logit $\hat S(\mathbf{c}_t)$. The action $a_t$ used to condition $\hat Q$ is the first action of the demonstrated chunk; the resulting scalar advantage is applied as a per-sample weight on the chunk-level flow-matching loss. Since $\hat Q$ is conditioned on $a_t$ while $\hat V$ marginalizes over the dataset action distribution at $\mathbf{c}_t$ via expectile regression, the advantage $\hat Q(\mathbf{c}_t, a_t) - \hat V(\mathbf{c}_t)$ remains informative even though both heads regress on the same Monte-Carlo target (see Appendix~\ref{app:rl-objectives}). Because training data consists exclusively of successful demonstrations, this advantage is weakly identified in a strict offline-RL sense and is more naturally interpreted as a \emph{trajectory-phase reweighting term} (Appendix~\ref{app:rl-objectives}).

\paragraph{Q loss.} We regress $\hat Q(\mathbf{c}_t, a_t)$ directly on the Monte-Carlo return-to-go $r_t$ with a Huber (smooth-$L_1$) penalty, avoiding bootstrapping and out-of-distribution action extrapolation:
\begin{equation}
\mathcal{L}_{\text{Q}} = \mathbb{E}_{(\mathbf{c}_t,a_t,r_t) \sim B}\left[ \mathrm{Huber}\left( \hat Q(\mathbf{c}_t,a_t) - r_t \right) \right],
\end{equation}
where $B$ denotes a mini-batch sampled from the training dataset.

\paragraph{Value loss with expectile regression.} We train $\hat V$ on the same Monte-Carlo target $r_t$ using IQL-style expectile regression~\cite{kostrikov_offline_2022}:
\begin{equation}
\mathcal{L}_{\text{V}} = \mathbb{E}_{(\mathbf{c}_t,r_t) \sim B}\left[ \rho \left(r_t-\hat V(\mathbf{c}_t)\right)^2_{+} + (1-\rho) \left(r_t-\hat V(\mathbf{c}_t)\right)^2_{-} \right],
\end{equation}
where $(\cdot)_{+}$ and $(\cdot)_{-}$ indicate positive and negative residuals, and $\rho \in (0.5, 1)$ (we use $\rho = 0.8$) biases $\hat V$ toward higher residuals, producing an optimistic baseline for in-dataset actions. The resulting advantage $A_t = \hat Q(\mathbf{c}_t, a_t) - \hat V(\mathbf{c}_t)$ is detached from the autograd graph and turned into a per-sample weight
\begin{equation}
w_{A,t} = \min \left\{ \exp(A_t /  \beta), C \right\},
\end{equation}
where $\beta$ and $C$ are the temperature and safety clip constants, respectively.

\paragraph{Success classification loss.} A binary success head $\hat S(\mathbf{c}_t)$ shares the value trunk and is trained with binary cross-entropy on a label that fires near task completion:
\begin{align}
\mathcal{L}_{\text{S}} &= \mathbb{E}_{(\mathbf{c}_t,a_t) \sim B} \!\left[ \mathrm{BCEWithLogits}\big( \hat S(\mathbf{c}_t), y_t^{\text{succ}} \big) \right], \\
y_t^{\text{succ}} &= \mathbb{I}\{ r_t \ge r_{\text{succ}} \},
\end{align}
where $r_{\text{succ}}$ is a constant threshold. The resulting calibrated probability $p_{\text{succ}}(\mathbf{c}_t) = \sigma\!\left(\hat S(\mathbf{c}_t)\right)$ defines a per-sample weight $w_{S,t} = 0.5 + 0.5\, p_{\text{succ}}(\mathbf{c}_t)$ that reweights the flow-matching loss while keeping all samples contributing.

\paragraph{Total training loss.} We train a flow-matching policy with the following total objective:
\begin{equation}
\label{eq:total-loss}
\begin{aligned}
\mathcal{L}_{\text{total}}
&= \mathbb{E}_{(\mathbf{c}_t, a_t) \sim B}\!\left[
w_{A,t} w_{S,t} \ell_{\text{FM}}(\mathbf{c}_t, a_t)
\right] \\
&\quad + \lambda_V \mathcal{L}_{V}
+ \lambda_Q \mathcal{L}_{Q}
+ \lambda_S \mathcal{L}_{S}.
\end{aligned}
\end{equation}
where $\ell_{\text{FM}}(\mathbf{c}_t, a_t)$ is the per-sample flow-matching loss whose expectation is $\mathcal{L}_{\text{FM}}$ in Eq.~\ref{eq:flow-matching-loss}, $w_{A,t}$ and $w_{S,t}$ are the advantage- and success-derived weights (both detached from the autograd graph), and $\lambda_V$, $\lambda_Q$, $\lambda_S$ control the relative contribution of the value, action-value, and success losses. The multiplicative weight $w_{A,t}\, w_{S,t}$ retains the simplicity of supervised flow matching while shifting probability mass toward contexts the critic associates with progress on successful trajectories, without requiring any on-policy rollouts.
\section{Experimental evaluation}
We evaluate ProgVLA on two standard simulation benchmarks and a real-robot setup, reporting the \emph{success rate (SR)}: the proportion of evaluation episodes that satisfy the task-specific success criterion. LIBERO~\cite{liu_libero_2023} defines four categories (\emph{Spatial}, \emph{Object}, \emph{Goal}, \emph{Long}) with ten tasks per category (40 tasks total); we use the official datasets (50 episodes per task, 2{,}000 total) and the official evaluation protocol. For Meta-World~\cite{yu_meta-world_2019}, we use the MT50 LeRobot dataset 
\footnote{\website{https://huggingface.co/datasets/lerobot/metaworld\_mt50}},
which contains 49 of the 50 nominal tasks in the public release, with 50 demonstrations per task.

\subsection{Implementation details}
We merge all tasks within a benchmark into a single dataset and train a separate model for each benchmark. Crucially, ProgVLA is trained \emph{without any large-scale robot pretraining}: all task-specific parameters (resamplers, fusion transformer, action expert, and progress heads) are initialized randomly and trained from scratch on the target benchmark data alone. The only pretrained components are the publicly available vision backbones (DUNE or DINOv3, both ViT-Small) and the frozen T5 text encoder, which provide generic visual and language priors but contain no robot data. The complete model with two image inputs has $109\text{M}$ parameters in total, of which $74\text{M}$ are updated during training.\footnote{For the Meta-World evaluation, the model has $100\text{M}$ parameters, of which $65\text{M}$ are updated during training, because the benchmark uses only a single-view camera input.} Full hyperparameters, training schedule, resampler token counts, and per-benchmark training times are provided in Appendix~\ref{app:details}.

\subsection{Simulation benchmark results}
\paragraph{LIBERO evaluation.} We follow the LIBERO evaluation protocol~\cite{liu_libero_2023}: training and evaluation use wrist and agent camera images, proprioception (end-effector pose and joint angles), and the natural-language instruction; each task is evaluated 50 times with random initial states. Baseline scores are imported from~\cite{shukor_smolvla_2025} and were not rerun under our exact protocol; the comparison should therefore be read as one \emph{against published results} on LIBERO rather than a strictly matched head-to-head evaluation, with the main protocol differences (demo counts and per-task trial counts) discussed in Appendix~\ref{app:baseline-protocol}.

\begin{table*}[ht]
\centering
\begin{tabular}{lccccc}
\toprule
\textbf{LIBERO benchmark} & \textbf{Spatial} & \textbf{Object} & \textbf{Goal} & \textbf{Long} & \textbf{Avg.} \\
\midrule
ProgVLA (0.1B) & \hlcell{third}87.6 & \hlcell{best}\textbf{96.0} & \hlcell{second}92.0 & \hlcell{best}\textbf{88.6} & \hlcell{best}\textbf{91.1} \\
\midrule
Octo (0.09B)~\cite{ghosh_octo_2024} & 78.9 & 85.7 & 84.6 & 51.1 & 75.1 \\
$\pi_{0}$ (Paligemma-3B)~\cite{black_pi_0_2024} & 87 & 63 & 89 & 48 & 71.8 \\
$\pi_{0}$ (3.3B)~\cite{black_pi_0_2024} & \hlcell{second}90 & 86 & \hlcell{best}95 & \hlcell{third}73 & \hlcell{third}86.0\\
OpenVLA (7B)~\cite{kim_openvla_2024} & 84.7 & 88.4 & 79.2 & 53.7 & 76.5 \\
SmolVLA (0.24B)~\cite{shukor_smolvla_2025} & 87 & \hlcell{third}93 & 88 & 63 & 82.75\\
SmolVLA (2.25B)~\cite{shukor_smolvla_2025} & \hlcell{best}93 & \hlcell{second}94 & \hlcell{third}91 & \hlcell{second}77 & \hlcell{second}88.75\\
\bottomrule

 \\
\toprule
\textbf{Meta-World benchmark} & \textbf{Easy} & \textbf{Medium} & \textbf{Hard} & \textbf{Very hard} & \textbf{Avg.} \\
\midrule
ProgVLA (0.1B) & \hlcell{third}84.9 & \hlcell{best}\textbf{72.7} &\hlcell{best} \textbf{77.0} & \hlcell{best}\textbf{79.6} & \hlcell{best}\textbf{78.5} \\
\midrule
$\pi_{0}$ (Paligemma-3B)~\cite{black_pi_0_2024} & 80.4  &  40.9  & 36.7  & 44.0  & 50.5 \\
$\pi_{0}$ (3.3B)~\cite{black_pi_0_2024} & 71.8 & \hlcell{third}48.2 & \hlcell{third}41.7 & 30.0 & 47.9\\
SmolVLA (0.24B)~\cite{shukor_smolvla_2025} & \hlcell{second}86.43 & 46.36 & 35 & \hlcell{third}60 & \hlcell{third}56.95\\
SmolVLA (2.25B)~\cite{shukor_smolvla_2025} & \hlcell{best}87.14 & \hlcell{second}51.82 & \hlcell{second}70 & \hlcell{second}64 & \hlcell{second}68.24\\
\bottomrule
\end{tabular}

\caption{Success rates (\%) on simulated LIBERO and Meta-World tasks. We report averages over 50 episodes per task; baseline scores are imported from~\cite{shukor_smolvla_2025,kim_openvla_2024}.}
\label{table:benchmark_result}
\end{table*}

ProgVLA (0.1B) attains the highest average success rate ($91.1\%$) across LIBERO task families, with the strongest performance on Object ($96.0\%$) and Long-horizon ($88.6\%$). It outperforms OpenVLA (7B) by $14.6$ points overall and by $34.9$ points on long-horizon tasks, and it exceeds SmolVLA at its largest scale (2.25B parameters) by $2.4$ points overall and by $11.6$ points on long-horizon tasks. Unlike $\pi_0$, OpenVLA, and SmolVLA, which leverage large-scale robot pretraining on cross-embodiment datasets, ProgVLA is trained from scratch on $2{,}000$ LIBERO trajectories, with only off-the-shelf DUNE and T5 components initialized from public pretrained weights. Subject to the protocol caveats above, a 0.1B-parameter model trained without robot pretraining outperforming substantially larger pretrained baselines indicates that the architecture delivers strong long-horizon performance with an order of magnitude fewer parameters and far less robot data than existing methods.

\paragraph{Meta-World evaluation.} Meta-World offers a simpler observation setting than LIBERO: a single corner-placed RGB camera and a 4-D proprioceptive vector (end-effector position and gripper state), following~\cite{shukor_smolvla_2025}. We examine multi-task generalization over the public MT50 LeRobot dataset, which currently contains 49 of the 50 nominal Meta-World tasks, grouped into four difficulty levels (easy, medium, hard, very hard) as defined in~\cite{seo_masked_2022}, with the same evaluation protocol as~\cite{shukor_smolvla_2025}. Baseline numbers are imported from~\cite{shukor_smolvla_2025} (Appendix~\ref{app:baseline-protocol}). ProgVLA matches the best baselines on easy tasks and substantially outperforms them on the harder tiers: $+20.9$ points on medium, $+7.0$ points on hard, and $+15.6$ points on very hard relative to SmolVLA (2.25B), yielding a $+10.3$-point overall improvement. Because ProgVLA is trained on MT50 alone without any robot-data pretraining while $\pi_0$ and SmolVLA benefit from large-scale robot pretraining, the gap on harder tiers is particularly informative: the architecture remains effective in challenging settings even without additional robot data.

\subsection{Ablation study}
To quantify the contribution of each component, we conduct an ablation study on LIBERO (Table~\ref{table:ablation_libero}). The four variants are: (1) \emph{w/o progress objectives} removes the progress heads and their losses (the imitation loss is no longer reweighted, and no auxiliary supervision is applied to $\mathbf{c}_t$); (2) \emph{w/o context resampler} feeds the fusion transformer's full output tokens directly to the flow-matching policy, removing the post-fusion bottleneck; (3) \emph{DINOv3} replaces the DUNE backbone with DINOv3 features\footnote{Input images are resized to $(128\times128)$ for DINOv3 to produce a number of patch tokens comparable to that generated by DUNE using $(112\times112)$ inputs.}; and (4) \emph{frozen DUNE} keeps the DUNE backbone fixed during training.

All four components contribute, but with markedly different magnitudes. The post-fusion context resampler (variant~(2)) is the single most critical component: removing it causes a $16.0$-point drop in average success rate and a particularly severe degradation on long-horizon tasks ($88.6 \to 51.2$). Fine-tuning the visual backbone is the second-largest lever: freezing DUNE (variant~(4)) reduces average success by $13.5$ points. The progress-aware auxiliary objectives (variant~(1)) yield a smaller but consistent $2.3$-point average drop when removed, with the gap concentrated on the Object ($96.0 \to 90.6$) and Long ($88.6 \to 85.1$) categories, consistent with the hypothesis that an internal progress signal is most useful when the policy must track multiple objects or maintain coherent behavior over many steps. Replacing DUNE with DINOv3 (variant~(3)) remains competitive overall ($88.7$ vs.\ $91.1$), with the gap again concentrated on long-horizon tasks. Per-variant discussion is provided in Appendix~\ref{app:ablation-detail}.

\begin{table*}[ht]
\centering
\begin{tabular}{lccccc}
\toprule
\textbf{Ablation} & \textbf{Spatial} & \textbf{Object} & \textbf{Goal} & \textbf{Long} & \textbf{Avg.} \\
\midrule
ProgVLA (0.1B) & 87.6 & 96.0 & 92.0 & 88.6 & 91.1 \\
\midrule
(1) w/o progress objectives & 87.0 & 90.6 & 90.2 & 85.1 & 88.8 \\
(2) w/o context resampler & 84.4 & 77.2 & 87.4 & 51.2 & 75.1 \\
(3) DINOv3 backbone & 88.2 & 92.4 & 93.2 & 81.0 & 88.7 \\
(4) frozen DUNE & 79.0 & 85.2 & 85.4 & 60.6 & 77.6 \\
\bottomrule
\end{tabular}
\caption{Ablation study on LIBERO. Numbers denote success rates (\%) averaged over 50 episodes per task.}
\label{table:ablation_libero}
\end{table*}

\subsection{Real-world experiment}
\begin{figure}
  \centering
  \includegraphics[width=7.5cm]{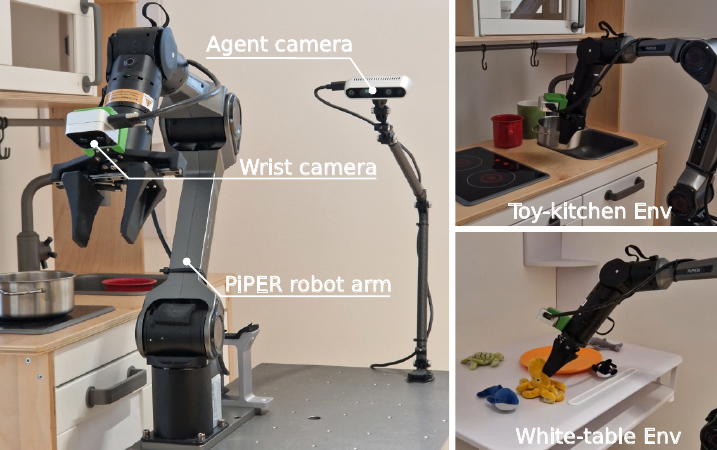}
  \caption{Real-world experimental setup: a 6-DOF PiPER arm with a gripper and two cameras.}
  \label{fig:real_robot}
\end{figure}

We train ProgVLA from scratch on data acquired on a real-world robotic platform: a 6-DOF PiPER arm with a gripper, a wrist camera (RealSense D405), and an agent camera (RealSense D435), shown in Fig.~\ref{fig:real_robot}. We define 10 manipulation tasks across two environments (a toy kitchen and a white-table setup), collect 50 teleoperated demonstrations per task, and train with the same hyperparameters as the simulation benchmarks (Appendix~\ref{app:details}). The trained policy runs in real time on a portable laptop (Intel Ultra 7 CPU, RTX 3500 GPU). Hardware, object lists, and observation-space details are provided in Appendix~\ref{app:realworld-setup}.

The trained model is evaluated in the same setups used during data collection, with initial environment configuration and robot pose randomly sampled to match the training distribution. Over 100 trials (10 per instruction across 10 tasks), ProgVLA achieves a $68\%$ average success rate; per-instruction success rates are reported in Table~\ref{tab:real_result} (Appendix~\ref{app:realworld-setup}). 
Among the 32 failed trials, the three most frequent failure modes are obstruction by scene clutter (10 cases), gripper-opening timeouts (8 cases), and grasping the wrong object on the white-table setup (4 cases); a detailed breakdown is provided in Appendix~\ref{app:failure-modes}. Because real-world training, evaluation, and data collection share the same two environments, these numbers establish in-distribution feasibility rather than cross-environment generalization. Overall, the results demonstrate that the same recipe used in simulation transfers to a physical robot without architectural changes.
\section{Conclusion and discussion}
\label{sec:conclusion}
We introduced ProgVLA, a compact 0.1B-parameter VLA combining two-stage Perceiver resampling, a universal DUNE vision backbone, and auxiliary progress-aware heads trained jointly with a flow-matching policy. On LIBERO and Meta-World, ProgVLA reaches success rates that exceed substantially larger pretrained baselines on multi-object, long-horizon, and high-difficulty tasks, despite using \emph{no} large-scale robot pretraining. Ablations isolate the learned context resampler and task-adaptive fine-tuning of the visual backbone as the dominant performance levers, while progress-aware auxiliary training contributes a smaller but consistent gain concentrated on long-horizon and multi-object tasks. The same recipe transfers without modification to a real 6-DOF arm across ten tasks. These results suggest that, in the compact regime, careful sequence compression and a strong universal visual prior can substitute for cross-embodiment robot pretraining, with internal progress supervision providing an additional boost where horizons grow.

Several directions remain open. Our evaluation covers simulated LIBERO and Meta-World and a single real robot with in-distribution evaluation; broader real-world testing across diverse platforms and held-out environments is needed to assess robustness. Our shaped target $r_t$ is purely temporal, so ``progress-aware'' here refers to the training \emph{objective} rather than a verified internal progress estimator; direct validation of the heads (calibration of $p_{\text{succ}}$, monotonicity / ranking analyses for $\hat V$) and replacing the temporal target with subgoal- or semantics-derived targets are natural next steps. Finally, the same heads could be extended with on-policy refinement, uncertainty estimation, or exploitation of failed trajectories. Appendix~\ref{app:limitations} provides a fuller discussion of limitations, deployment safety, and broader impact.
\section{Acknowledgment}
This research was supported in part by the Ministry of Trade, Industry and Energy (MOTIE, Republic of Korea) under the project “Development of human-robot shared autonomy-based robot intelligence enhancement technology and RaaS Business Model for logistics robots” (Project No. 00445855), supervised by the Korea Evaluation Institute of Industrial Technology (KEIT).
{
    \small
    \bibliographystyle{ieeenat_fullname}
    \bibliography{zotero}
}

\clearpage

\appendix
\section{Notation summary}
\label{app:notation}

For convenience we collect the notation used throughout the paper in Table~\ref{tab:notation}.

\begin{table*}[ht]
\centering
\small
\begin{tabular}{ll}
\toprule
\textbf{Symbol} & \textbf{Meaning} \\
\midrule
$t$ & Discrete time-step index within an episode.\\
$T_n$ & Terminal time index of episode $n$ (assumed a success state).\\
$T_M$ & Fixed horizon cap used to normalize the progress target ($T_M=500$).\\
$\mathbf{I}_t = [I^1_t,\ldots,I^C_t]$ & RGB images from the $C$ cameras at step $t$.\\
$\mathbf{q}_t$ & Proprioceptive vector (joint angles, joint velocities, gripper status).\\
$l_n$ & Natural-language instruction for the $n^{\text{th}}$ episode.\\
$\mathbf{A}_t = [a_t,\ldots,a_{t+H-1}]$ & Action chunk of horizon $H$ starting at step $t$ (prediction horizon).\\
$H_e$ & Execution horizon (number of actions actually executed before replanning).\\
$\mathbf{c}_t$ & Set of context tokens output by the multi-modal encoder.\\
$\pi_\phi$ & Multi-modal encoder (parameters $\phi$).\\
$\pi_\theta$ & Flow-matching action expert (parameters $\theta$).\\
$\hat Q(\mathbf{c}_t, a_t)$ & State-action critic head; regresses $r_t$ on $(\mathbf{c}_t,a_t)$.\\
$\hat V(\mathbf{c}_t)$ & State value head; expectile regression on $r_t$ from $\mathbf{c}_t$ alone.\\
$\hat S(\mathbf{c}_t)$ & Success logit head; binary head on $y_t^{\text{succ}}$.\\
$r_t \in [0,1]$ & Shaped temporal progress target $\max(0, 1 - (T_n-t)/T_M)$.\\
$r_{\text{succ}}$ & Threshold on $r_t$ that defines the success label $y_t^{\text{succ}}$.\\
$A_t$ & Advantage $\hat Q(\mathbf{c}_t,a_t) - \hat V(\mathbf{c}_t)$, detached from autograd graph.\\
$w_{A,t}$, $w_{S,t}$ & Detached per-sample weights on the imitation loss derived from $A_t$ and $p_{\text{succ}}$.\\
$\beta$, $C$ & Temperature and clip in $w_{A,t} = \min\{\exp(A_t/\beta), C\}$ ($\beta=0.5$, $C=20$).\\
$\rho$ & Expectile parameter in $\mathcal{L}_V$ ($\rho=0.8$).\\
$\lambda_V$, $\lambda_Q$, $\lambda_S$ & Loss weights on $\mathcal{L}_V$, $\mathcal{L}_Q$, $\mathcal{L}_S$ ($0.5$, $1.0$, $0.2$).\\
\bottomrule
\end{tabular}
\caption{Notation used in the main text.}
\label{tab:notation}
\end{table*}

\section{Additional experimental details}
\label{app:details}

\subsection{Justification of the offline-RL-inspired objectives}
\label{app:rl-objectives}

\paragraph{Q loss design.}
A standard offline-RL critic is trained with a TD target $r + \gamma\,\hat V(\mathbf{c}')$, which requires bootstrapping and is well known to suffer from overestimation bias on out-of-distribution actions~\cite{kumar_stabilizing_2019}. Since we work entirely from successful demonstrations and use $r_t$ as a shaped progress signal rather than a sparse reward, we deliberately replace the TD target with the \emph{Monte-Carlo return-to-go} $r_t$ itself. This eliminates bootstrapping, target networks, and OOD action extrapolation at the cost of giving up the Bellman backup. We choose a Huber (smooth-$L_1$) penalty for robustness, as it attenuates heavy-tailed noise and outliers common in long-horizon trajectories, improving numerical stability compared to $L_2$ while preserving smooth gradients. Although $\hat Q$ and $\hat V$ share the same regression target, they remain distinct functions: $\hat Q$ is conditioned on the specific dataset action $a_t$ via the concatenated action token, while $\hat V$ summarizes only the context tokens.

\paragraph{Value loss design.}
We borrow the expectile-regression trick from Implicit Q-Learning (IQL)~\cite{kostrikov_offline_2022} but apply it to the same Monte-Carlo target $r_t$ used by $\mathcal{L}_{\text{Q}}$, rather than to a bootstrapped Bellman target. This sidesteps value extrapolation on out-of-distribution actions entirely, while still letting us recover IQL's key property: an asymmetric value estimate that upweights positive residuals and therefore behaves like an \emph{optimistic} baseline over in-dataset actions. With $\rho > 0.5$ (we use $\rho = 0.8$), $\hat V$ is biased toward the higher residuals attainable at $\mathbf{c}_t$, so that dataset actions whose return-to-go exceeds the action-marginalized estimate produce a positive advantage. The symmetric Huber regression of $\hat Q$ tracks the conditional return for the demonstrated action $a_t$, while the asymmetric expectile regression of $\hat V$ tracks an optimistic envelope of returns at the same context. The advantage is detached from the autograd graph before being used to reweight the imitation loss, so $\mathcal{L}_{\text{V}}$ and $\mathcal{L}_{\text{Q}}$ act as auxiliary supervision rather than direct policy-gradient signals.

\paragraph{Success classification rationale.}
By supervising the model with an easy-to-learn binary success signal, this auxiliary task stabilizes and enriches the learned representations and produces a calibrated success probability $p_{\text{succ}}(\mathbf{c}_t) = \sigma(\hat S(\mathbf{c}_t))$ used to reweight the flow-matching loss. The reweighting $w_S = 0.5 + 0.5\, p_{\text{succ}}(\mathbf{c}_t)$ leaves all samples contributing while emphasizing those close to task completion, improving sample efficiency and credit assignment near the critical end-of-task segments where small action errors most often determine success.

\paragraph{On the identifiability of the advantage and the action-granularity choice.}
Because our training data consists exclusively of successful demonstrations, each context $\mathbf{c}_t$ is effectively paired with a single logged action. In this regime, the difference $\hat Q(\mathbf{c}_t,a_t) - \hat V(\mathbf{c}_t)$ is only weakly identified as an action-quality signal in the strict offline-RL sense, and is more naturally interpreted as a \emph{trajectory-phase reweighting term} that emphasizes contexts at which the demonstrated trajectory is closer to completion. We deliberately do not claim a stronger offline-RL interpretation. The action $a_t$ used to condition $\hat Q$ is the first action of the demonstrated chunk $\mathbf{A}_t = [a_t, \ldots, a_{t+H-1}]$, and the resulting scalar advantage is applied as a per-sample weight on the chunk-level flow-matching loss. Using a single anchor action keeps the critic lightweight, while the multiplicative coupling on the chunk-level imitation loss preserves the structure of supervised flow matching.

\paragraph{Why the multiplicative weights help long-horizon tasks.}
Inspecting Eq.~\ref{eq:total-loss}, the imitation loss for each $(\mathbf{c}_t,a_t)$ is scaled by the product $w_{A,t}\, w_{S,t} \in [0.5\,\exp(\text{clipped advantage}/\beta),\, C]$. Two effects follow. First, because $r_t$ is increasing in trajectory phase, $\hat V$ and $\hat Q$ are typically larger near the end of a trajectory, and so are their differences; the expectile bias in $\hat V$ further sharpens $A_t$ near phase changes (e.g., between an approach phase and a grasp phase) where the return-to-go distribution at a given context is asymmetric. Second, $w_{S,t}$ explicitly upweights frames close to task completion. On long-horizon tasks, the most fragile decisions are often clustered near the end of a trajectory (precise grasps, final placements, sub-task transitions); the multiplicative weights direct extra capacity to exactly these decisions, which qualitatively matches our ablation finding that the largest gain from the progress heads appears on the Object and Long categories of LIBERO. We did not perform additional analyses to verify these mechanisms directly, and leave such targeted probes to future work.

\subsection{Architectural design considerations}
\label{app:design-considerations}

\paragraph{Why two stages of resampling.} A single fusion bottleneck applied to raw modality tokens has two issues. First, the relative number of tokens contributed by vision, language, and proprioception is highly imbalanced (hundreds of patch tokens from a ViT vs.\ a few dozen language tokens vs.\ a single proprioception token), and a one-stage Perceiver tends to allocate most of its latent capacity to the modality with the largest input. Second, a single bottleneck has to simultaneously balance modalities and prepare control-ready tokens, which entangles two distinct functions. Our two-stage design separates these concerns: per-modality resamplers normalize the number of tokens emitted by each encoder, while the post-fusion resampler produces the small set of context tokens consumed by the action expert and the progress heads. This is also what makes adding the progress heads cheap: they only see the post-fusion tokens, so their cost scales with the (small) number of context tokens rather than the (large) number of raw patch tokens.

\paragraph{Why fine-tune the vision backbone.} Both DUNE and DINOv3 are trained on general images that are visually quite different from typical robot wrist-camera views (which are often close-up, motion-blurred, and have unusual viewpoints). The ablation in Table~\ref{table:ablation_libero} shows a $13.5$-point drop when DUNE is frozen, with most of the loss again concentrated on long-horizon tasks. We interpret this as evidence that fine-tuning the backbone is needed to specialize attention to wrist-camera viewpoints, fine-grained gripper/object contact regions, and the lighting conditions present in the simulator and real setups.

\paragraph{Why a shared trunk for $\hat V$ and $\hat S$.} The value head and the success head both consume only the context tokens $\mathbf{c}_t$ and both regress quantities that depend on trajectory phase. Sharing a Perceiver pooler between them encourages the pooled representation to be informative about \emph{both} continuous progress and the binary near-completion event, while keeping the additional parameters introduced by the auxiliary objectives small relative to the action expert. The action-value head $\hat Q$ uses a separate Perceiver pooler because it additionally consumes the action token, and we did not want the action conditioning to leak into the inputs of $\hat V$ and $\hat S$ via a shared trunk.

\paragraph{Alternatives we did not pursue.} We considered three alternatives during development. (i) A behavior-cloning baseline without any progress signal, which corresponds to ablation variant (1) in Table~\ref{table:ablation_libero}; this is $2.3$ points worse on average. (ii) Using $r_t$ directly as a per-sample loss weight (i.e., $w_t = r_t$), bypassing $\hat Q$ and $\hat V$; we found this less effective in early experiments because $r_t$ is purely an open-loop function of trajectory phase and does not adapt to the difficulty of a particular context, but a full comparison was not run for the final model. (iii) A subgoal-derived target instead of the temporal $r_t$; we did not pursue this because reliable subgoal annotations are not available across LIBERO and Meta-World, and we wanted a recipe that requires no extra annotation beyond the existing demonstrations. We flag (iii) as a natural extension in Sec.~\ref{sec:conclusion}.

\subsection{Training hyperparameters and schedule}
\label{app:training}

All models are trained for 500 epochs with a batch size of 256. We optimize using AdamW ($\beta_1 = 0.95, \beta_2 = 0.999$) with an initial learning rate of $1\times10^{-4}$. The learning rate is warmed up for 500 steps and then decayed using a cosine schedule down to $1\times10^{-6}$. To stabilize optimization, we apply gradient norm clipping with a maximum norm of $1.0$ and maintain an exponential moving average (EMA) of the model parameters with decay $0.75$.

For the visual backbone, we employ DUNE~\cite{sariyildiz_dune_2025} and DINOv3~\cite{simeoni_dinov3_2025}, using the small-scale variants of their publicly available pretrained models built on the ViT-Small architecture. The input images are resized to $(112\times112)$ for DUNE and to $(128\times128)$ for DINOv3 (to roughly match the number of patch tokens). The vision backbone is shared across all camera inputs, while a separate Perceiver Resampler is trained for each camera.

All Perceiver Resamplers in the model (per-camera, language, and the post-fusion cross-modal resampler, as well as the pooling resamplers inside $\hat Q$ and $\hat V$) use the same architecture: 2 stacked cross-attention blocks with hidden width $d_{\text{model}}=384$, $4$ attention heads inside the head-internal poolers and $8$ heads inside the encoder-side resamplers, MLP expansion ratio $4$, and dropout $0.1$. The vision Perceiver Resampler produces $8$ tokens per image, the language Perceiver Resampler produces $16$ tokens, and the final cross-modal Resampler yields a contextual representation comprising $4$ tokens. The shared cross-modal fusion Transformer between the per-modality and post-fusion resamplers is a $2$-layer Pre-LN Transformer encoder with $d_{\text{model}}=384$, $8$ attention heads, MLP expansion ratio $4$, and dropout $0.1$.

The action expert follows the SmolVLA-style interleaved cross-/self-attention design: $12$ blocks alternating cross-attention to the post-fusion context tokens and self-attention over the action sequence, with $d_{\text{model}}=384$, $8$ attention heads per block, MLP expansion ratio $4$, and a continuous sinusoidal time embedding mixed in via AdaLN-Zero modulation. During training, the flow-matching time variable is drawn from $\tau \sim \mathrm{Beta}(\alpha=2,\,\beta=2)$, following~\cite{shukor_smolvla_2025}. At inference, we integrate the learned flow ODE with $K=10$ Heun (explicit trapezoidal) steps; we found Heun to be slightly more accurate than pure Euler at the same step count without measurable latency impact in our setting.

The action expert is trained to predict action sequences in chunks of length $16$ (prediction horizon). At inference time, we execute only the first $8$ steps of each predicted chunk (execution horizon). With the exception of the T5 text encoder~\cite{raffel_exploring_2020}, which remains frozen, all remaining parameters are trained jointly, including the DUNE vision backbone, the per-modality and cross-modal resamplers, the fusion Transformer, the progress heads, and the action expert.

For the training objective in Eq.~\ref{eq:total-loss}, we set the loss weights to $\lambda_V = 0.5$, $\lambda_Q = 1.0$, and $\lambda_S = 0.2$. The advantage reweighting uses temperature $\beta = 0.5$ and clip $C = 20.0$ in $w_{A,t} = \min\{\exp(A_t/\beta),\, C\}$, with the advantage detached from the autograd graph. The expectile in the value loss is $\rho = 0.8$. The success label threshold is $r_{\text{succ}} = 1 - 17/T_M \approx 0.966$ with $T_M = 500$, so $\hat S$ is supervised to fire only on roughly the final $17$ steps of each demonstration.

A single NVIDIA H100 GPU was used for training. The LIBERO model required a total training time of $25$ hours, and the Meta-World model required $32$ hours. The real-world model is trained for approximately $8.5$ hours on the same hardware. Code, pretrained checkpoints, and the collected real-world dataset will be released upon publication.

All numbers reported in the main text are from single training seeds for each configuration; during development we observed approximately $\pm 1\text{--}2$ success points of run-to-run variance across LIBERO categories. We did not perform a sensitivity sweep over $T_M$, $\rho$, $\beta$, or $C$, and we leave such analyses, together with multi-seed evaluation and a direct calibration study of the progress heads, to future work.

This is a deliberately restrictive setup that lets us isolate the contribution of the proposed architecture and training objectives; it should be kept in mind when comparing absolute numbers to baselines (such as $\pi_0$, OpenVLA, and SmolVLA variants) that rely on extensive robot pretraining on cross-embodiment corpora (e.g., Open X-Embodiment, DROID).

\subsection{Detailed ablation discussion}
\label{app:ablation-detail}

Per-variant analysis of the ablation results in Table~\ref{table:ablation_libero}:

\paragraph{(1) Removing the progress objectives.} Removing the progress heads and their losses yields a moderate drop in average success rate ($91.1 \to 88.8$), indicating that progress-aware supervision provides a useful learning signal but is not solely responsible for the gains. The drop is most consistent on the Object ($96.0 \to 90.6$) and Long-horizon ($88.6 \to 85.1$) categories, suggesting that progress-aware training is particularly helpful when the policy must track multiple objects or maintain coherent behavior over longer horizons. This is consistent with the mechanism described in Appendix~\ref{app:rl-objectives}: the multiplicative weights $w_{A,t}\,w_{S,t}$ direct gradient signal toward the end-of-trajectory frames where precise grasps and placements are most consequential.

\paragraph{(2) Removing the post-fusion context resampler.} This leads to a substantial degradation, especially on long-horizon tasks ($88.6 \to 51.2$), and lowers the average performance by $16$ points. This highlights the central role of the post-fusion bottleneck in fusing and distilling visual observations, language instructions, and proprioception into a small set of informative tokens. Without this bottleneck, the action expert must attend to a much larger and less structured set of tokens, which appears to harm long-horizon behavior disproportionately, possibly because the increased context length makes the cross-attention layers in the action expert more sensitive to spurious correlations.

\paragraph{(3) Changing the visual backbone from DUNE to DINOv3.} Overall performance is close to the full model ($88.7$ vs.\ $91.1$), but with a noticeable drop on long-horizon tasks ($88.6 \to 81.0$). This indicates that while a generic backbone such as DINOv3 is already competitive, using DUNE as the visual backbone still yields an additional performance margin, particularly when temporal context must be maintained. We hypothesize that DUNE's multi-source distillation (which includes geometric and depth-style targets) provides priors that are more directly useful for manipulation than purely contrastive image features.

\paragraph{(4) Freezing DUNE during training.} This produces a $13.5$-point reduction in average success rate ($91.1 \to 77.6$) and a large decrease on long-horizon tasks ($88.6 \to 60.6$), underscoring the importance of jointly optimizing the visual representations and the policy. The gap is consistent with the view that off-the-shelf vision backbones, however strong, are not specialized to wrist-camera viewpoints or fine-grained gripper/object contact regions, and need targeted adaptation during policy training.

Taken together, these trends indicate that (i) the post-fusion context resampler is the single most critical architectural component for robust long-horizon performance, (ii) fine-tuning the visual backbone on the target benchmark provides the second-largest gain, and (iii) progress-aware auxiliary training contributes a smaller but consistent additional gain concentrated on long-horizon and multi-object tasks.

\subsection{Real-world experimental setup}
\label{app:realworld-setup}

\paragraph{Hardware.} The setup uses two low-cost PiPER robot arms from Agile-X\footnote{\url{https://global.agilex.ai/products/piper}}. One arm serves as the teleoperation leader (used only for data collection), while the other acts as the follower during data collection and is subsequently used as the execution robot for real-world evaluation.

\paragraph{Observation and action spaces.} Visual observations are RGB images first resized and then center-cropped to a final resolution of $(112\times112)$ pixels. The proprioceptive state includes the end-effector pose, joint angles, and gripper status. The action space is defined by the joint-angle displacements of the arm and the gripper angle.

\paragraph{Environments and objects.} Datasets for eight tasks (the top eight in Table~\ref{tab:real_result}) are collected in the toy kitchen environment shown in Fig.~\ref{fig:real_robot}. The objects present in this environment are a pot, a green mug, a red mug, a toy octopus, and a toy blue whale. The remaining two tasks are conducted in the white-table environment, which uses a set of five soft ocean-animal toys: a blue whale, a gray whale, a yellow octopus, a turtle, and a dolphin.

\paragraph{Per-instruction results.} Table~\ref{tab:real_result} reports the per-instruction success rates over $10$ trials per task. Per-task success ranges from $50\%$ (close the left cabinet door under the sink; pick up the yellow octopus and place it in the pot) to $80\%$ (three pick-and-place instructions involving the kitchen sink), with an average of $68\%$ across all 100 trials.

\begin{table*}
    \centering
    \begin{tabular}{lc}
    \toprule
    \textbf{Instruction} & \textbf{SR} \\
    \midrule
    close the left cabinet door under the sink & 50\\
    close the right cabinet door under the sink & 70\\
    pick up the blue whale and place it in the pot & 70\\
    pick up the green cup and place it in the kitchen sink & 80\\
    pick up the pot and place it in the kitchen sink & 80\\
    pick up the pot and place it on the front stove & 60\\
    pick up the red cup and place it in the kitchen sink & 80\\
    pick up the yellow octopus and place it in the pot & 50\\
    \midrule
    pick up the blue whale and place it on the orange plate & 70\\
    pick up the yellow octopus and place it on the orange plate & 70\\
    \midrule
    Average & 68\\
    \bottomrule
    \end{tabular}
    \caption{Language instructions and their corresponding evaluation results in a real-world toy kitchen environment (top) and white-table environment (bottom).}
    \label{tab:real_result}
\end{table*}

\subsection{Real-world failure-mode analysis}
\label{app:failure-modes}

Of the 32 failed trials in the real-world evaluation, three failure modes dominate.

The most frequent cause is obstruction by other objects in the scene. For example, when instructed to pick up a red cup and place it in the sink, the motion sometimes fails because a green cup or a pot blocks the arm's path; such obstacle-related issues account for $10$ of the $32$ failures.

In $8$ cases, the robot fails to open the gripper within the allotted time, leading to a timeout. This typically occurs when the gripper has closed firmly on an object and the policy issues an opening command, but the actual joint motion does not complete before the trial timer expires.

In $4$ cases on the white-table setup, the robot grasps a different animal from the one specified in the instruction. The remaining failures arise from a mix of less frequent issues, including imprecise final placement and occasional collisions with the table edge.

These failure modes are largely interpretable and suggest concrete avenues for improvement, such as obstacle-aware path planning, longer per-trial time budgets, and additional language-grounding supervision for fine-grained object discrimination. We additionally note that none of the failure modes are specific to the progress heads themselves; the heads are consumed only at training time, and the deployed model is a single flow-matching policy whose error modes are therefore similar to those expected of any imitation-learned manipulation controller.

\subsection{Baseline-protocol differences}
\label{app:baseline-protocol}
The baseline numbers in Table~\ref{table:benchmark_result} are imported from~\cite{shukor_smolvla_2025} rather than rerun under our exact protocol, and there are two protocol differences that the reader should keep in mind. First, on LIBERO, ProgVLA is trained on the official LIBERO datasets with $50$ demonstrations per task ($2{,}000$ demonstrations in total), whereas the SmolVLA baselines in~\cite{shukor_smolvla_2025} are reported on a filtered LIBERO release containing $1{,}693$ episodes after removing failed trajectories and no-op actions. Thus, ProgVLA is exposed to a larger number of demonstrations, but these demonstrations come from the original LIBERO release with lower-resolution images, while the imported baselines use a smaller but filtered release with regenerated higher-resolution images. Second, ProgVLA is evaluated for $50$ episodes per task on LIBERO, while some imported numbers are reported with $10$ episodes per task, so ProgVLA's per-task success rates have lower evaluation noise than the imported numbers. On Meta-World, the public MT50 LeRobot release currently contains $49$ of the $50$ nominal tasks (one task is missing from the public release at the time of writing), so the ``MT50'' label is used loosely. We did not rerun the baselines under our exact protocol due to compute constraints, and therefore the headline numbers in Table~\ref{table:benchmark_result} should be interpreted as a comparison \emph{against published results} on these benchmarks rather than as a strictly matched head-to-head evaluation.

Two specific consequences of these differences are worth flagging. (i) The image-resolution gap goes against ProgVLA: the imported SmolVLA numbers benefit from higher-resolution renderings of LIBERO, while ProgVLA uses the original, lower-resolution release. (ii) The demonstration-count gap goes in ProgVLA's favor: $2{,}000$ vs.\ $1{,}693$ demonstrations is a roughly $18\%$ increase. We do not believe either effect changes the qualitative picture: ProgVLA's gap over SmolVLA-2.25B on long-horizon LIBERO ($11.6$ points) is far larger than the typical effect of an $18\%$ data increase reported in the SmolVLA paper, and ProgVLA's $10$+ point gap on the harder Meta-World tiers cannot be explained by either of the LIBERO-specific differences. A strictly matched re-run of the baselines under our exact protocol is left to future work.

\subsection{Extended limitations, broader impact, and outlook}
\label{app:limitations}

\paragraph{Extended limitations.}
Beyond the summary in Sec.~\ref{sec:conclusion}, there are a few extra points worth highlighting. (i) Our evidence centers on policy success rate; targeted validation of the progress heads (calibration curves for $p_{\text{succ}}$, monotonicity / ranking analyses for $\hat V$ along trajectories, separation between successful and failed rollouts) is an important next step. (ii) The shaped target $r_t = \max(0,\, 1 - (T-t)/T_M)$ is purely a function of remaining trajectory length and a fixed cap $T_M$; episodes that exceed the cap or contain qualitatively distinct phases of the same length are treated identically, and we did not sweep $T_M$. (iii) The auxiliary objectives use only successful demonstrations, so the resulting advantage signal is weakly identified in a strict offline-RL sense; failed trajectories are not exploited. (iv) The real-robot evaluation is restricted to a single 6-DOF arm with $10$ tasks, where training and evaluation share the two environments used for data collection; this establishes in-distribution feasibility and partial robustness but not cross-platform or cross-environment generalization. (v) All numbers come from single training runs; during development we observed approximately $\pm 1$ success points of run-to-run variance across LIBERO categories, so small ablation differences should be interpreted as consistent trends rather than precisely calibrated effects. (vi) Although we report total/trainable parameter counts and training times, we did not include a detailed efficiency study (inference latency, control rate, peak memory, or FLOPs); the model runs in real time on a portable laptop in our real-world experiments, but a systematic efficiency benchmark is left to future work. (vii) We did not perform a sensitivity sweep over $T_M$, $\rho$, $\beta$, or $C$, or a separated ablation of the $Q$, $V$, and $S$ heads.

\paragraph{Broader impact and safety.}
A compact VLA that delivers strong long-horizon performance without large-scale robot pretraining can lower the barrier to entry for robot learning in academic and small-lab settings, which we view as a positive societal effect: it widens the set of institutions that can study and develop language-conditioned robot policies without access to cross-embodiment pretraining infrastructure. At the same time, deployment of imitation-learned manipulation policies on physical hardware raises the usual safety considerations: our real-world failure analysis (Appendix~\ref{app:failure-modes}) shows that wrong-object grasps, scene-clutter obstructions, and gripper-opening timeouts can occur, and we recommend that any deployment include human supervision, hardware safety interlocks, and an explicit task and object scope rather than assuming open-world reliability. We also note that, like other imitation-learned policies, ProgVLA inherits any biases present in its demonstration data (preferred grasps, preferred object orderings); these biases should be audited before any safety-critical deployment.

\paragraph{Outlook.}
Natural next steps include direct validation of the progress signal (calibration and ranking analyses for $\hat V$ and $p_{\text{succ}}$), multi-seed evaluation, sensitivity studies on $T_M$, $\rho$, $\beta$, and $C$, separated ablations across the auxiliary heads, broader real-world evaluations with held-out objects and environments, on-policy refinement that allows the policy to benefit from rollouts (and from failed trajectories), and integration of richer modalities (force/torque, touch, audio) and language-driven planning to broaden the tasks addressable by compact, progress-aware VLAs. Another promising direction is replacing the purely temporal progress target with a learned or semantically-derived target (e.g., from a vision-language progress estimator such as~\cite{ma_vision_2025} or from human-annotated subgoals), which would relax the assumption that all dataset trajectories are equally well-paced and could yield a more semantically meaningful progress signal at no architectural cost.

\end{document}